\documentclass[
  journal=medium,
  manuscript=article-type,
  year=2023,
  volume=37,
]{cup-journal}
\usepackage{blindtext}
\usepackage{booktabs}
\usepackage{url}
\usepackage{setspace}
\expandafter\def\expandafter\UrlBreaks\expandafter{\UrlBreaks
  \do\a\do\b\do\c\do\d\do\e\do\f\do\g\do\h\do\i\do\j%
  \do\k\do\l\do\m\do\n\do\o\do\p\do\q\do\r\do\s\do\t%
  \do\u\do\v\do\w\do\x\do\y\do\z\do\A\do\B\do\C\do\D%
  \do\E\do\F\do\G\do\H\do\I\do\J\do\K\do\L\do\M\do\N%
  \do\O\do\P\do\Q\do\R\do\S\do\T\do\U\do\V\do\W\do\X%
  \do\Y\do\Z\do\1\do\2\do\3\do\4\do\5\do\6\do\7\do\8\do\9\do\0}

\title{Towards Human-Level Text Coding with LLMs: The Case of Fatherhood Roles in Public Policy Documents}

\author{Lorenzo Lupo}
\affiliation{Computing Sciences Department, Bocconi University, Via Sarfatti 25, 20136 Milan, Italy.}
\email[Lorenzo Lupo]{lorenzo.lupo2@unibocconi.it}

\author{Oscar Magnusson}
\affiliation{Department of Political Science, University of Gothenburg, Sprängkullsgatan 19, 41123 Göteborg, Sweden.}

\author{Dirk Hovy}
\affiliation{Computing Sciences Department, Bocconi University, Via Sarfatti 25, 20136 Milan, Italy.}

\author{Elin Naurin}
\affiliation{Department of Political Science, University of Gothenburg, Sprängkullsgatan 19, 41123 Göteborg, Sweden.}

\author{Lena Wängnerud}
\affiliation{Department of Political Science, University of Gothenburg, Sprängkullsgatan 19, 41123 Göteborg, Sweden.}


\begin{document}

\maketitle

\begin{abstract}
Recent advances in large language models (LLMs) like GPT-3.5 and GPT-4 promise automation with better results and less programming, opening up new opportunities for text analysis in political science.
In this study, we evaluate LLMs on three original coding tasks involving typical complexities encountered in political science settings: a non-English language, legal and political jargon, and complex labels based on abstract constructs.
Along the paper, we propose a practical workflow to optimize the choice of the model and the prompt.
We find that the best prompting strategy consists of providing the LLMs with a detailed codebook, as the one provided to human coders. In this setting, an LLM can be as good as or possibly better than a human annotator while being much faster, considerably cheaper, and much easier to scale to large amounts of text.
We also provide a comparison of GPT and popular open-source LLMs, discussing the trade-offs in the model's choice.
Our software allows LLMs to be easily used as annotators and is publicly available at 
\url{https://github.com/lorelupo/pappa}.
\\
\end{abstract}

\section{Introduction}
\label{sec:intro}
Political science is among the social sciences that have most eagerly embraced automatic text analysis methods \citep{grimmer2021machine}. 
Indeed, in many downstream research applications, such as data collection, topic classification, and especially content analysis, political scientists have found natural language processing (NLP) techniques to hold significant potential in terms of cost-saving and scaling capacity \citep{slapin2008scaling,grimmer2010bayesian,roberts2016model}. 
Yet, despite their promise, data scarcity (the lack of sufficient high-quality training data and the cost and time associated with creating them) and a lack of technical expertise have thus far prevented NLP methods from being widely implemented---both in political science and the social sciences more generally. 

Recently, a new generation of large language models (LLMs) has changed the cost-efficiency calculus for political scientists interested in the possibilities of NLP techniques. LLMs, including the renowned GPT series, can not only produce coherent texts almost indistinguishable from those of human writers, but they can also automate many tasks---including coding text for analysis.
This is possible because the model has a general concept of language and only needs to be instructed on how to solve the task at hand,
a significant improvement over both traditional statistical text classification approaches and the most modern transfer learning ones \citep{laurer_van2023_trans}, which still require thousands of hand-coded data points.
In addition, APIs like OpenAI's or the Hugging Face Hub allow the use of LLMs with little programming knowledge.
These properties make LLMs attractive for political scientists who work with text but might not have the resources to manually annotate thousands of examples and the computational knowledge to run supervised approaches.

In this paper, we use a concrete political science project to study text annotation with LLMs.
Specifically, we define three different, interdependent coding tasks of varying difficulty about ``fatherhood roles'' in Swedish political discourse (see Section \ref{sec:study} for details). 
While several research projects have already shown the LLM's prowess in text annotation, few - if any - have investigated their performance in a typical political science setting involving a minority language (like Swedish) with legal or political jargon and complex labels (see \citet{ollion2023chatgpt} for an overview).
Indeed, the different fatherhood roles we define are theoretical constructs with subtle differences from one another, making the coding task harder than many typical NLP coding tasks like sentiment analysis, where human annotators typically have a high intercoder agreement.
Moreover, the novelty of our annotation task offers a robust test of the LLMs' generalizability, as the possibility that the LLM has memorized the task during its training (data contamination) can be ruled out.
Finally, evaluating LLMs on complex tasks where a "ground truth" does not exist requires benchmarking against multiple human annotators. One element stands out in our contribution, compared to the existing literature: we acknowledge this lack of consensus by assessing the LLMs against not one, but three trained political scientists.
In our work, we evaluate the performance of two renowned GPT models (GPT-3 and GPT-4) and three Open-Source (OS) LLMs, experimenting with some relevant choices in the LLM-coding methodology (Section \ref{sec:methodology:prompteng}): joint versus disjoint coding of multiple interdependent tasks, level of detail in the prompt description of labels, zero-shot versus few-shot learning, and the order of coding examples provided in few-shot learning.    
For each of these settings, we compare the performance of the language models to that of three human coders along the various tasks.

Our main \textbf{findings} are: 1) LLMs annotation is optimal when they are treated as human coders: providing them with an exhaustive codebook of label definitions and several coding examples; 2) Following this prompting strategy, some LLMs perform on par or outperform human coders, even in complex annotation tasks involving a minority language, linguistic specificity, and complex labels; 3) The popular GPT-4 model is at the frontier of annotation performance in the most difficult task, while OS LLMs catch up on the easier tasks; 4) Jointly performing three coding tasks on the same texts results in comparable performance to performing them separately (approximately three times cheaper and faster).

\section{The Example Case Study}
\label{sec:study}
As a case study for the proposed methodology, we investigate the occurrence of different predefined father roles in Swedish policy documents. The task is derived from a real-world research question on changing societal norms in Sweden from a forthcoming paper and incurs a large degree of interpretation on the annotators as it relies on complex abstract constructs, that were defined specifically for this research question. In this sense, the task is theoretically challenging, and the low agreement between human annotators (see \ref{sec:eval}) indicates that annotators disagree on the "right" label in more than just a few cases. 
So far, the literature has evaluated LLMs on coding tasks that were either theoretically simpler (such as the political polarity of a sentence), or based on English texts, where current LLMs are typically the strongest, or both (see \citet{ollion2023chatgpt} for an overview). Our task is representative of many cases encountered in political science research, involving complex labels and minority languages.
Moreover, the novelty of out coding task guarantees that our evaluation is not subject to ``data contamination'' \citep{golchin2023time} since it is improbable that the LLMs have encountered the tasks we are evaluating during their training.
Contamination arises when models are trained on popular, open-source text datasets that may include the very tasks on which they are evaluated, and potentially even the specific text examples that are present in their test set.
Such exposure during training leads to models essentially memorizing correct responses, resulting in inflated performances on the ``contaminated'' test sets.
This phenomenon affects popular datasets across various scientific fields, including political science, casting doubt on the reported performance of models evaluated on them.
While current LLMs might have read the Swedish documents collected in our dataset during their training phase, the coding tasks are surely novel to them and cannot have been memorized.


\subsection{Data}\label{sec:methodology:data}
We used publicly available data from the Swedish parliament (Riksdagen)\footnote{https://www.riksdagen.se/sv/dokument-och-lagar/riksdagens-oppna-data/} including  government investigations, propositions, motions, and other legislative texts. 
We collected all public documents from 1993 to 2021, converted them to raw text format, and lowercased them.
We selected only sentences referencing fathers (i.e., containing the Swedish words pappa, pappor, fäder, fader, far).
We added further checks to ensure the term actually referred to a father: for example, the word ``far'' can also mean ``go'' or ``travel''  if used as a verb.
We used a part-of-speech tagger to filter out those instances.
The final data set contains 1,910 sentences.
Data collection and filtering details are reported in \ref{app:corpus}.

\section{Methodology}\label{sec:methodology}


\subsection{Coding paradigm}\label{sec:methodology:labels}

After an initial qualitative analysis of the dataset, the theoretically derived labels were applied to a small (16) sample of the data. We classified each of the Swedish texts on fatherhood along three tasks, each represented by a varying number of suitable labels. The three labeling tasks target different aspects of the texts (content vs.\ form), are interconnected, and hierarchical. If the first task is ``not applicable'', the other two tasks are irrelevant and thus not coded. \ref{app:corpus} presents a detailed description of each label.

\paragraph{Task 1 - Type of paternal involvement} in the family. Possible labels for this task are ``passive,'' ``active negative,'' ``active positive caring,'' ``active positive daring,'' ``active positive other,'' and ``not applicable.''

\paragraph{Task 2 - Explicitness of the description}, denoting the linguistic means by which the father's role is expressed in the sentence. Possible labels: ``implicit'' and ``explicit.''

\paragraph{Task 3 - Normativeness of the description}, denoting whether the sentence expresses a factum or a desired outcome. Possible labels: ``descriptive'' and ``ideal.''

\subsection{Validation set}\label{sec:methodology:validset}

A validation set was created (350 sentences) to establish a baseline of human performance and provides a direct comparison set for the LLMs' coding choices. Measuring LLMs performance allows one to optimize the prompt provided to them and to choose the most suitable LLM. 
To build a validation set, researchers should provide human annotators with a codebook describing the task and the labels, and providing some labelled examples.
Three authors independently coded the same random subset of 350 sentences following our codebook, which can be found in \ref{app:corpus:codebook}. 

\subsection{Prompt}\label{sec:methodology:prompt}
Similarly to humans, LLMs need to be instructed to annotate text. In our experiments we evaluated different prompting strategies, showing that the best way to prompt LLMs is by providing them with the same codebook that was built for humans. Here we describe its main components: the instruction, the description of the labels, and some labelled examples.
The codebook usually starts with an \textbf{instruction} defining the task. The instruction we used for all of our three coding tasks reads as follows:\footnote{Note that we used English instructions, even though the text coded by the model is in Swedish. In fact, in preliminary experiments, we found that Swedish instructions alongside Swedish texts resulted in inferior performances compared to English instructions.}
\begin{quote}
\textit{Label the Swedish text according to how it describes the role of the father in the family. Possible labels are:}
\end{quote}

The list of labels defined in the \textit{coding paradigm} step (Section \ref{sec:methodology:labels}), should then follow. Each label should also be accompanied by an exhaustive \textbf{description} of its meaning. In our case, the labels of the first task needed longer descriptions, as the definitions of fatherhood roles are more abstract and complex than the labels of the other two tasks.
\textbf{Labelled examples} should also be provided to the coders, especially highlighting cases of ambiguity. We randomly sampled example sentences from our corpus (excluding the validation set) and annotated them. We discussed and agreed on the labels assigned to the examples.

\subsection{Language Models}\label{sec:methodology:model}

Many options exist today when it comes to LLMs, both open source (OS) and proprietary, like the GPTs. Every LLM has distinct costs and performances on different tasks and different languages. For a new coding task, performances are hard to know in advance. Evaluating the most promising LLMs on the validation set allows to for an informed choice.
We experimented with two of the best proprietary LLMs to date, i.e., the GPT-3 model \texttt{text-da-vinci-003} and GPT-4 \citep{openai2023gpt4}
from OpenAI\footnote{via their API \url{https://platform.openai.com/docs/api-reference/introduction}}, and two of the best OS models: Llama-2 (the \texttt{-13b-chat-hf} version; \cite{touvron2023llama}) and Mixtral-8x7b (the \texttt{-Instruct-v0.1} version. Alongside, we also tested a less popular OS model which was fine-tuned on Nordic languages: GPT-SW3 (the \texttt{-20b-instruct} version; \cite{ekgren2023gptsw3}).

\subsection{Prompt engineering}\label{sec:methodology:prompteng}

To optimize LLMs' coding performance, different prompting combinations can to be evaluated on the validation set. In our case, we experimented with the following elements of the codebook:

\paragraph{Tasks} 
Indicates whether the model is prompted to code a single task at a time or the three tasks simultaneously for each text that is provided in input. In the former case, the model outputs a single label for each text, while in the latter it generates three, one for each task.
For example, in Figure \ref{fig:main1}, we prompted the model to code all the three tasks together in the column "multi-task".
The motivation is to allow the model to use shared information in the hierarchical coding scheme and to make the coding process more efficient, as the model does not have to be run three times on the same instance to provide a label for each task.

\paragraph{Label descriptions} Indicates different degrees of explanation of the labels provided to the model. From no description, to a short description, to a long and comprehensive one (as the labels' description provided in the codebook). We experimented with all the three variants to code task 1 (type of paternal involvement), where labels have complex semantics. In contrast, we coded the second (explicitness) and third (normativeness) tasks with self-explanatory labels. 
Further explanation is in these cases redundant and hence not provided to the model.

\paragraph{Number of example sentences} 
We experimented with both few-shot and zero-shot learning. In the former case, we provided 15 labeled Swedish text examples to the model after the instructions and the definition of the labels. In the zero-shot case, we provided none.

\paragraph{Examples order} In \ref{app:recency}, we investigate how different orders of the example sentences provided in the prompt affect coding quality. Although the effect is not significant, we suggest strategies to avoid ``unfavorable'' orders.

\section{Empirical Results}
\label{sec:eval}

\begin{figure}[t!]
\centering
\includegraphics[width=\linewidth]{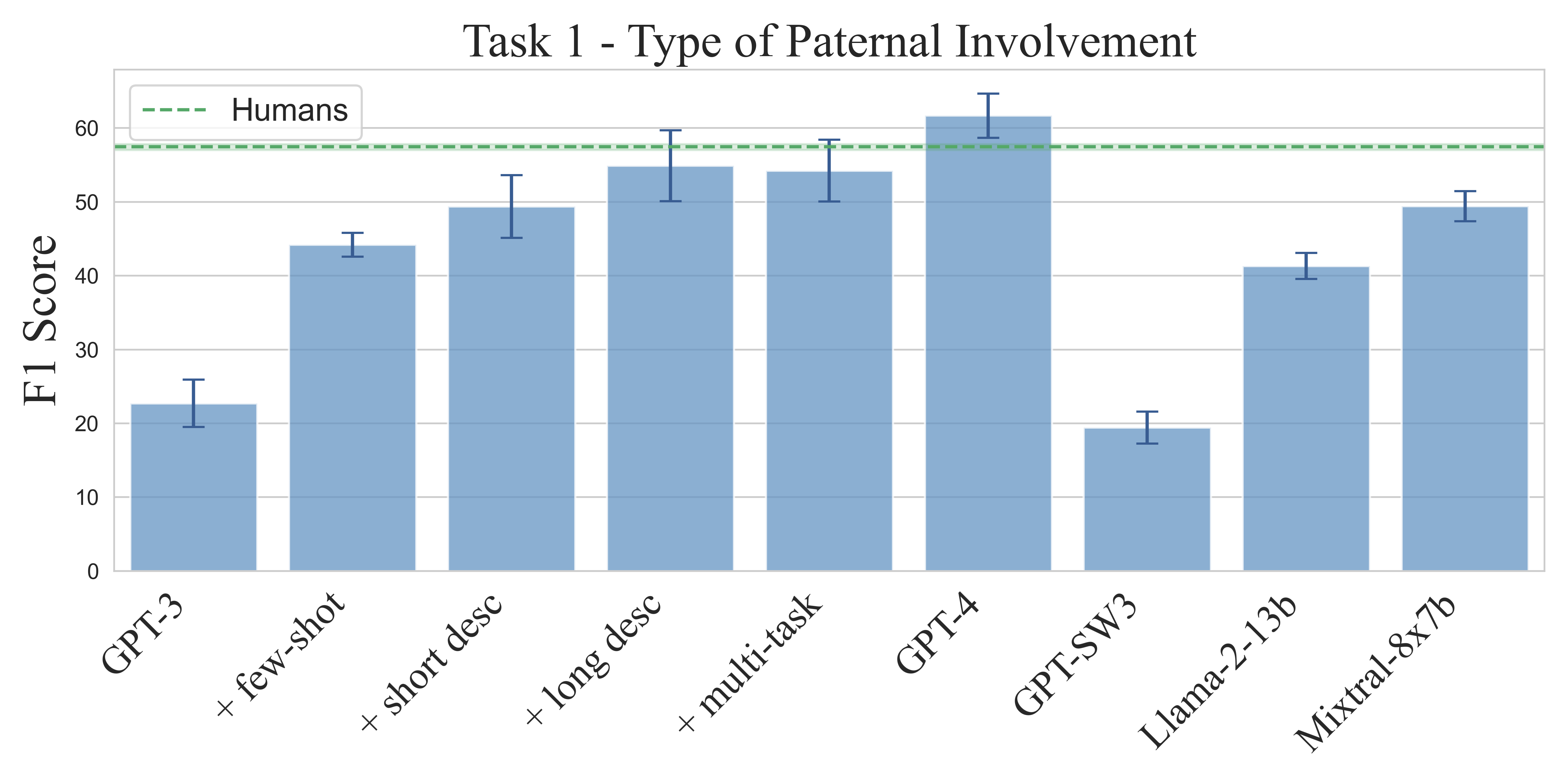}
\caption{Average coding agreement with (other) human coders on task 1: type of paternal involvement (6 labels).}\label{fig:main1}
\end{figure}

\begin{figure}[t!]
\centering
\includegraphics[width=\linewidth]{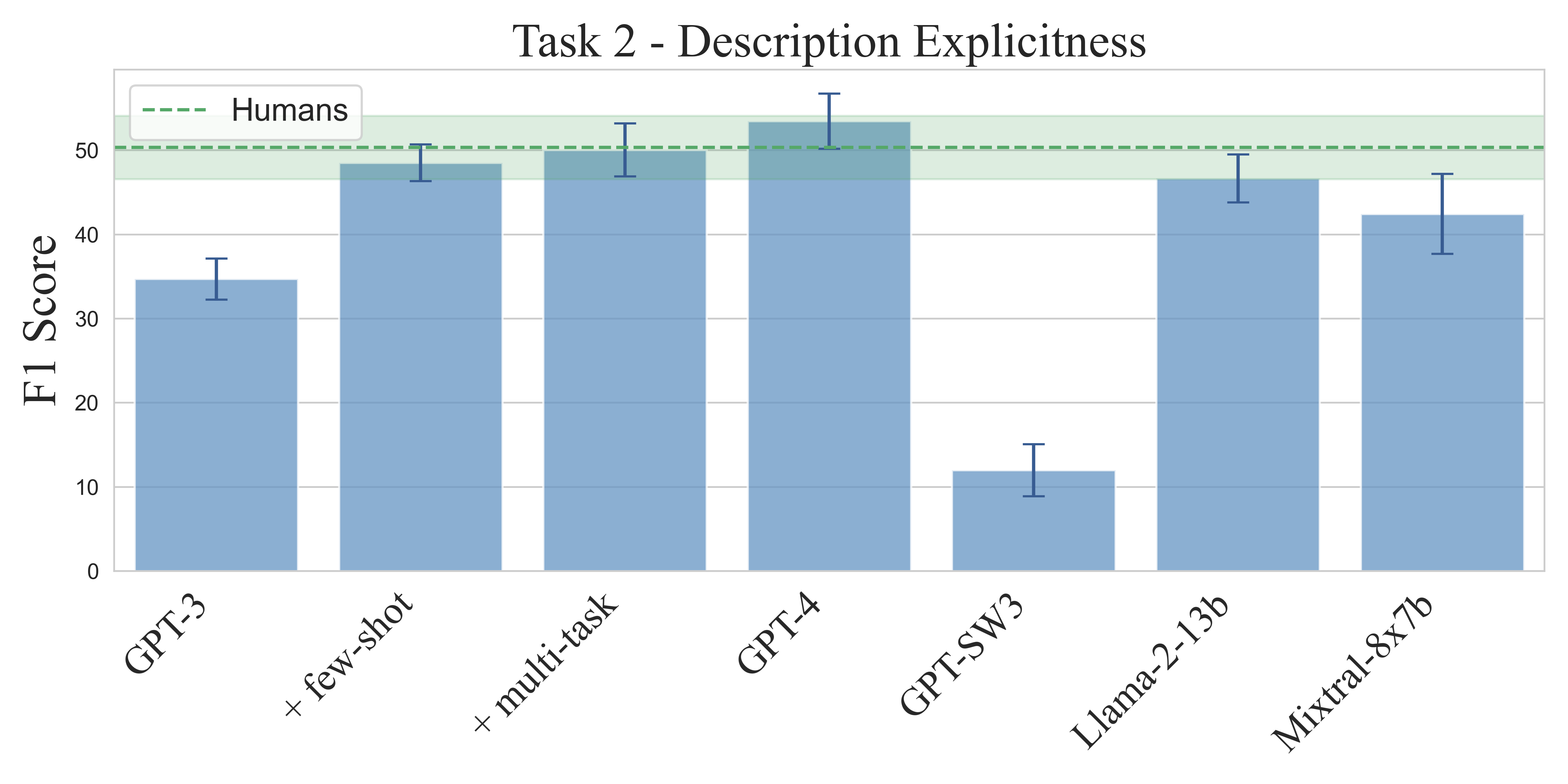}
\caption{Average coding agreement with (other) human coders on task 2: explicitness of the description (2 labels).}\label{fig:main2}
\end{figure}

\begin{figure}[t!]
\centering
\includegraphics[width=\linewidth]{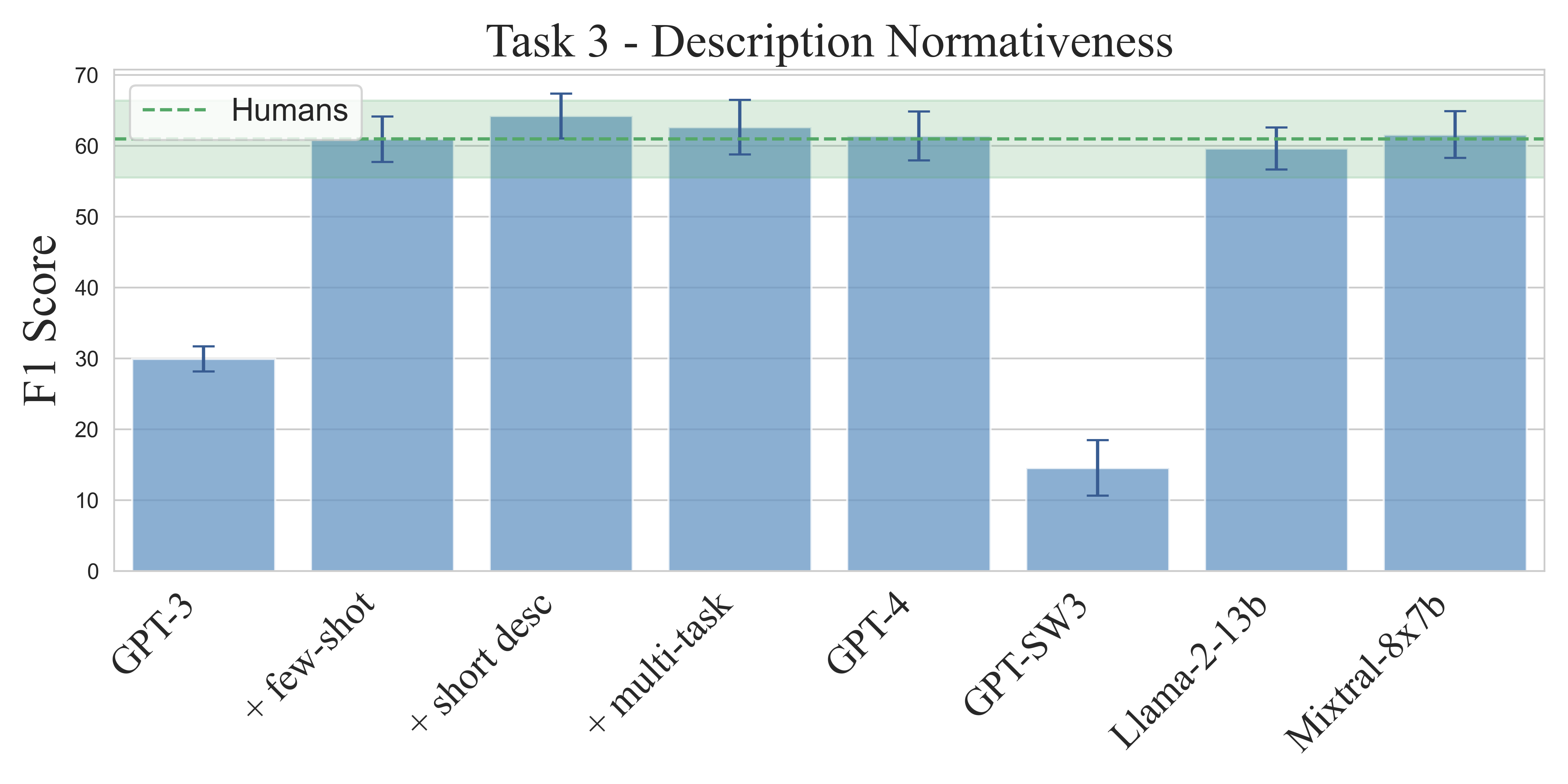}
\caption{Average coding agreement with (other) human coders on task 3: normativeness of the description (2 labels).}\label{fig:main3}
\end{figure}

To investigate humans inter-reliability, we compare the predictive performance (macro-F1) of each of the three human coders against the labels of the other two. We then plot the humans' scores average (green dashed line) and standard deviation (green shade) in Figures \ref{fig:main1}, \ref{fig:main2}, and \ref{fig:main3} for tasks 1, 2, and 3, respectively. In the same Figures we plot the performance average and standard deviation of each prompt configuration and model evaluated against humans' labels.

The average F1 score for humans ranges between 60 and 50 for the tasks. 
Less-than-perfect results for the human coders do not necessarily signal bad training or bad coders but usually indicate how difficult a task is, and the absence of a ground truth. The takeaway is that annotating fatherhood roles is tricky.

\subsubsection{Prompt Engineering}

The first bars on the left of each figure represent the performance of GPT-3 with different prompting strategies. In the first bar, GPT-3 is prompted to annotate tests in a zero-shot learning (zero labelled examples) and without any description of the labels beside their list. This prompt results in bad performance across the board.
Switching to few-shot learning (second bar in each Figure) significantly improves performances, even in the absence of any description of the labels.
Including labels' descriptions in relevant tasks (tasks 1 and 3; the labels for task 2 are self-explanatory) further improves results, especially when descriptions are exhaustive, i.e., long descriptions for the task 1, short descriptions for task 3.
The combination of few-shot learning and comprehensive label definitions results in the best performance, comparable to that of human coders.
In other words, GPT-3 performs best when prompted with the same codebook that is provided to human coders.

Interestingly, GPT-3 performs similarly when labeling the three tasks jointly (multi-task) and separately, in different runs.
Performance on the second task is slightly better with joint labeling, while the contrary is true on the third task.
This is good news because we only need to prompt the model to label each text once instead of three times - reducing the time and monetary costs to almost a third.

\subsection{Examples Order}

\begin{table}[t]
\begin{tabular}{cccc}
\toprule
\textbf{Examples order} & \textbf{Kappa} & \textbf{Raw} & \textbf{F1} \\
\midrule
0 & \textbf{49.36} & \textbf{62.57} & \textbf{54.92} \\
1 & 48.27 & 61.90 & 53.51 \\
2 & 46.58 & 60.67 & 53.72 \\
\bottomrule
\end{tabular}
\caption{GPT-3's average agreement with human coders on task 1, for three different random orders of the 15 labelled sentences provided in the prompt as examples.}\label{tab:exampleorder}
\end{table}

Not only does the number of examples we provide in the prompt affect the output of a language model, but also the order of the examples, because LLMs are affected by a `recency bias'' \citep{zhao2021calibrate}.
Here, we study its effect on labelling performance. 
Table~\ref{tab:exampleorder} shows the results of three different GPT-3 runs over the validation set, where the prompt contains a different \textit{random} order of the 15 examples provided in each run.
We only evaluate the most challenging task (Task 1: ``type of paternal involvement''). We prompted the model with only the label list for this task and long, comprehensive descriptions.
As expected from previous findings on recency bias, the performance changes with the order of the examples. 
Since we know which order performs best from trying different orders on the validation set, we can use this order within all other experiments involving few-shot learning.

\subsection{Other Models}

We then evaluate a mix of the most powerful models available to date using the best prompting strategy identified above (providing the full codebook).

Noticeably, GPT-4 is on par with human annotators across the board. The fact that some LLMs agree with human annotators as much as humans agree between themselves (or even more) might appear counter-intuitive.
However, we should remember that the models are coding the texts based on 15 examples whose labels \textit{all the human annotators agreed upon}.
Therefore, the models' coding is heavily influenced by coding choices that are shared by annotators and reflect this agreement.

OS LLMs (GPT-SW3, Llama-2-13b and Mixtral-8x7b) underperform the proprietary (GPT) models on the first task.
Interestingly, however, Llama-2-13b and Mixtral-8x7b catch up significantly on tasks 2 and 3, which are simpler (binary classification with less complex labels), becoming a viable alternative. 

\paragraph{} Our results show that the optimal coding choice for our tasks involves prompting the model with an exhaustive codebook, as the one provided to human annotators, and using GPT-4 in the multi-task setting or, in case of large datasets where GPT costs would be excessive, an OS LLM, which can be very competitive.
For more details and evaluation metrics (raw agreement and Cohen's $\kappa$) on our experiments, please refer to the Appendices.
Clearly, the performance evaluated on our validation set may not be perfectly generalizable to the corpus as a whole.
Ideally, subsequent manual evaluations could be conducted on a distinct set of sentences randomly selected from the rest of the corpus after its automatic annotation, performed with the optimal setup identified on the validation set.

\section{Conclusion and Outlook}

Many previous findings suggest that LLMs have great potential as annotators in the Political Sciences  \citep{ollion2023chatgpt}. In this paper, we contribute to that strand of research by showing that, if prompted as human annotators, both proprietary and open-source LLMs can perform on par with humans even when the annotation involves non-English texts, legalese and political jargon, and complex labels based on abstract constructs.
This complex setup, which is one encountered in many annotation projects for social science research, characterizes our first contribution with respect to the concurrent literature.
Second, we show that, like human annotators, LLMs annotate best when provided with a detailed codebook that includes clear instructions, an exhaustive definition of the labels, and some labeled examples covering a variety of cases.

It may seem provocative that LLMs' labels can have a higher quality than humans'. Most times, we treat human performance as the ground truth and the ability to mimic this performance as a measure of quality. As is the case in this paper. But as the far-from-perfect agreement between the humans make evident: humans make errors. While the errors may be small in relation to the closest coder, they might grow when comparing to the second closest, resulting in a subpar agreement on average. Of course, the LLMs make the same kind of errors, but it seems to distribute these errors so that it is closer to as many coders as possible, especially when all coders agree on the labelling of the few-shot examples provided to the model.  In this way, the LLM becomes a more reliable mirror image of the average human annotation, than that of other humans.

Meanwhile, LLMs are much faster and cheaper than human coders. We estimated that, as of November 2023, GPT-3 could code our full corpus 270 times faster than what humans would, saving 60\% of the money.\footnote{For details on this estimate, see \ref{app:speed}.}
Our empirical results suggest that some OS LLMs are a viable alternative to proprietary models like GPT for very large corpora, which would incur intolerable costs even with proprietary models. In particular, we showed that OS LLMs can perform as well as the GPTs on less theoretically challenging tasks. In the near future, however, the performance of OS LLMs is likely to approach that of current proprietary models in every task.

LLMs will likely have an outsized impact on political science methodology, not just in text analysis but also in text generation, such as survey questions, use of agent-based design, and manuscript outlines. 
The technology is here to stay, and we hope this paper helps practitioners to make sense of LLMs and adopt them responsibly and effectively for their needs. The software for evaluating and performing the annotation tasks of this paper was developed for seamless integration into other text coding projects. The script is publicly available at \url{https://github.com/lorelupo/pappa}. 




\paragraph*{Funding.} This work was supported by Riksbankens Jubileumsfond, project number RRD10-1418:1; Knut and Alice Wallenberg Foundation, project number KAW 2017.0245.

\paragraph*{Acknowledgments.} 
We would like to thank our colleagues at the Bocconi University's MilaNLP Lab for their constructive feedback on several versions of the manuscript.

\paragraph*{Data Availability Statement.}
Replication code and cleaned data are available on GitHub: \url{https://github.com/lorelupo/pappa}. 

\paragraph*{Competing interests:} 
The authors declare none.

\bigskip

\printbibliography
\appendix

\clearpage
\begin{center}\Large\bfseries
Appendices
\end{center}

\section{The evolution of language models}
OpenAI's GPT is one of the most advanced language models available to the public. Like all its current competitors, it builds probabilistic and neural language models. It is, therefore, useful to understand both families of models and how they work to get the most out of them.
This section represents a relatively technical introductory overview of these topics.

\subsection{Probabilistic Language Models}
Language models have been around for several decades. In their simplest form, they are a function that assigns a probability to any sentence $S$. That probability, $P(S)$ can be understood as the likelihood of encountering the sentence ``in the wild.''
Language models have been used to assess the output of various applications, for example, machine translation, and to re-rank the output based on their likelihood.

The most common way of estimating a probability is via maximum likelihood estimation (MLE)---or counting and dividing on a training corpus. The resulting probabilities can then be used (in theory) on any future input.
Here, we would count how often we have observed the sentence in a training corpus and divide it by the number of all sentences in that corpus.
One obvious obstacle to this approach is that there are countably infinite possible sentences (any sentence could be extended by adding another sentence via ``and''). The normalization factor is, therefore, infinity. Since dividing any number by infinity gives the answer 0, this approach does not work. Thus, language models have to be able to assign a probability to unseen sentences.

For the longest time, language models were based on the maximum likelihood estimation of $n$-grams. That is, they computed the probability of a sentence as the joint probability of all its words: $P(S) = P(w_1, w_2, ..., w_n)$. Again, the joint probability is hard to estimate directly. However, it can be broken down further into a sequence of conditional probabilities of word combinations (the $n$-grams). 

So, instead of computing the probability of ``I like platypuses,'', a 2-gram language model computes the probability of \textit{I} at the start of a sentence, the probability of \textit{like} given that the previous word was \textit{I}, and the probability of \textit{platypuses} given that the previous word was \textit{like}. Or
$$P(I, like, platypusses) = P (I|\texttt{START}) \cdot P(like|I) \cdot P(platypusses|like)$$
The first two can be computed by taking a large enough corpus and counting how often we see \textit{I} at the beginning of a sentence, and dividing by the count of all words starting a sentence ($P(I|\texttt{START})$). We can then count the number of times we observe \textit{I like} and divide by the count of any \textit{I} followed by any word.
The last word, \textit{platypuses}, is unlikely to occur in a corpus. Probabilistic language models use smoothing to ensure that a non-zero probability is assigned to every word in the vocabulary, even the rarest. The simplest form is Laplace smoothing, which adds a constant to the count of any word. To still produce a valid probability, we must recompute the normalization factor. There are various ways to do this.

Probabilistic language models of this type can be implemented in a few lines of code and assign reasonable probability estimates. Increasing the size of the $n$-grams produces better estimates. Still, it requires larger and larger corpora and more sophisticated smoothing \citep{chen1999empirical}.

Instead of scoring sentence probabilities, these models could also be used to generate new sentences by sampling from the learned conditional probability distributions. I.e., for each word, we select one from the most likely ones we have observed in that position.
The resulting sentences are locally coherent (i.e., any three words in a row would make sense) but often lack cohesion as they get longer and have no coherent meaning across sentence boundaries.

\subsection{Neural Language Models}
Probabilistic language models were limited by their reliance on $n$-grams and the Markov horizon to compute the necessary statistics. Larger $n$-gram windows produced better models but required larger corpora and more storage and improved only incrementally.
In the 2010s, a resurgence in the use of neural networks allowed for a dramatic leap forward in language models. Neural networks had been around since the 1950s \citep{rosenblatt1958perceptron}, but interest in them died off in mainstream AI as they were too inconsistent in their results. A variety of factors led to their resurgence.

The main factor is the availability of unprecedented amounts of data, driven by the popularization of the world wide web and, more recently, mobile applications. Neural networks have large numbers of parameters, and to set these parameters correctly, they need to be trained on large amounts of data. When IBM's Watson beat the leading Jeopardy! players by a wide margin in 2011, it used a data dump that comprised much of the collected knowledge available on the Internet from the year before. But the underlying technique, having independent algorithms come up with solutions and then weigh and combine them, was 20 years old. The technique had always been viable, but not without the vast amount of data that had only recently become available.
Indeed, a comparison of the most common AI problems shows that the algorithm that ``solved'' each of them was typically invented 18 years before the problem was solved.\footnote{\url{https://www.kdnuggets.com/2016/05/datasets-over-algorithms.html}} 
Meanwhile, the data sets that led to the breakthroughs were, on average, published just three years before.

Concurrently, researchers have made progress in understanding how these parameters should be computed and controlled, reducing the variability in neural network results. The development of better initialization regimes \citep{glorot2010understanding}, better non-linear activation functions \citep{glorot2011deep}, and an improved understanding of exploding or vanishing gradients \citep{hochreiter1991untersuchungen} has made it possible to train neural networks more efficiently and reliably.

The final factor that led to the resurgence of neural networks was computing power. Moore's law \citep{schaller1997moore} predicted doubling the number of transistors on a microchip every two years. While that relation has slowed, the computing power available on a simple smartphone today is magnitudes larger than anything available at the dawn of AI and even as recently as 20 years ago. Computations that would have taken years to finish on a specialized supercomputer can now be run on a standard laptop. Specialized hardware like graphical processing units (GPUs) has become even more specialized and is the architectural backbone of today's neural network models.

Neural networks, despite their names, are essentially universal function approximators. They can learn to replicate any function, provided they have enough data. This includes probability functions. Neural networks can directly compute the probability distribution over all words for any sequence without breaking it down into a sequence of conditional probabilities or limiting the horizon. This has made it possible to produce much longer coherent sentences with them.

\subsection{The Transformer Architecture and GPT}
Until recently, neural networks struggled with long-range dependencies, such as discourse phenomena in text sequences.
In the sentence ``This is \textbf{not}, by any stretch of the imagination, a \textbf{funny} movie,'' the keywords \textit{not} and \textit{funny} are far apart, yet need to be connected to make sense of the whole. In 2017, \citet{vaswani2017attention} introduced a groundbreaking neural architecture called Transformer that surpassed previous architectures used for sequence modeling, such as recurrent \citep{hochreiter1997long,chung2014empirical} and convolutional \citep{kim-2014-convolutional,dauphin2017language} neural networks.
At the heart of the Transformer architecture is the attention mechanism \citep{bahdanau2014neural}, a neural module allowing the elements of a sequence to be linked directly to one another, even when distant.
Transformer-based language models are today state-of-the-art and have gained public attention thanks to the remarkable linguistic prowess displayed by GPT models.

Until recently, neural networks excelled in specific tasks, such as sentiment analysis, but they couldn't bridge the gap with human-like versatility.
This gap is narrowing with the latest LLMs like GPT, which demonstrate a broader skill set, including software development, translation, mathematical derivations, and rhyming.


\section{The corpus and codes}\label{app:corpus}

Policy documents here refer to official texts written in everyday work by or for the Swedish Riksdag and Government. Our corpus includes various policy documents, for example all the motions, propositions, written questions, and interpellations published during the period 1993--2021, further described below.  The policy documents have in common that they all somehow preceded parliamentary decisions and include arguments for why a law should or should not pass. For instance, the Swedish lawmaking process is preceded by so-called public investigations via the SOU system (Statens offentliga utredningar). Going back in time, all laws have a corresponding ``SOU: year: number'' where skilled investigators provide the lawmaker with arguments for laws. It is likely that motions pertaining to the agentic role of the father that actually have had an impact on lawmaking (and therefore possibly on the behavior of men) would be found here. Furthermore, it is likely that government propositions and motions introduced by political parties and individual MPs will contain references to the role ascribed to fathers by lawmakers, i.e. parties who currently constitute the government as well as those in the opposition in Parliament.

We downloaded the policy documents from the website \url{https://data.riksdagen.se/data/dokument/}, which makes Swedish Riksdag and Government documents available as a publicly accessible dataset.

The Swedish names for the documents we include are: betänkanden, departementsserien, interpellationer, motioner, propositioner, protokoll, skriftliga frågor, statens offentliga utredningar (SOU), utredningar, utskottsdokument, yttranden, övrigt. We downloaded 16,390 documents under these headlines dating back to 1961. Most of them were motions (7,361), protokoll (3,263), betänkanden (1,549), propositions (1,244), and SOUs (1,116). Of these, we used only documents from 1993 and onward in this paper, as we cannot be sure that the documents available online for earlier periods represent complete sets of all documents of each type in those periods. 

The documents were processed in the following way: we downloaded all available documents, linked the content to the file ID and type, and extracted the data from PDFs where necessary. We lightly preprocessed the data (lowercasing and replacing numbers), then split it automatically into sentences, separating words and punctuation marks from each other. 
Thereafter, we identified target words for fathers, mothers and parents.

There were 277,168 sentences that included any of these keywords. We singled out the instances where these words were used as nouns, and not as verbs or other types of words. This means that our analyses take into account, for example, that ``far'' (father when used as a noun) is also a verb meaning ``go'' or ``travel.''  This left us with 151,489 sentences that include the keywords as nouns. The overwhelming proportion of these sentences talk about parents and do not single out fathers or mothers as separate actors: 1,901 sentences mention fathers and 2,821 mention mothers. 

The aim of our annotation is to detect the role that the father plays in the life of the family.
We divided the annotation into the labels described in the following codebook.

\subsection{Codebook}\label{app:corpus:codebook}

\paragraph{PASSIVE}
This is a category that captures what is sometimes called the classical role given to and taken by fathers, which in essence is a role that does not include engaging with the child in person.  Words used in theory to describe this role might be ``custodial parenthood,'' ``provider,'' ``breadwinner,'' or ``process parenthood.'' This is a person who is important in the life of the child, but is not involved in their hands-on care and upbringing. Society includes norms and sometimes creates clarifying policies around this father, for his sake, as well as for the sake of the child and the mother. Envisioned indicators might be words denoting the father being the moneymaker or provider, or someone who provides genes or sperm or overall protection or security.

\paragraph{ACTIVE NEGATIVE}
This is a category that captures the negative roles of fathers, such as the oppressive father, the harsh and violent father, as well as the neglectful father. Society needs to create policies around this father in order to restrict him. Envisioned indicators are notions such as dangerous, punish (not educate or discipline in a non-aggressive way), aggressive, violent, oppressive, or not paying child support.

We also have three categories that capture the positive roles of fathers. Society needs to create policies around this father in order to encourage him. These categories are:  

\paragraph{ACTIVE POSITIVE CARING}
This category focuses on what is stereotypically seen as the female parental role. Envisioned indicators would be words such as caring, warm, nurture, understanding, empathy, listening, comfort, or confirming

\paragraph{ACTIVE POSITIVE DARING}
This category focuses on what is stereotypically seen as the male parental role. Envisioned indicators would be things like strengthening activity, risk-taking, daring, test your limits, sport, outgoing, education (including reading to the child, which is one of the recurring activities in the data). 

\paragraph{ACTIVE POSITIVE OTHER}
There are also other ways of being active as a father, and this category captures that. Most typically, this category indicates that the father is competent and reliable, and should be trusted with the child, but it does not single out why. Envisioned indicators are general references to competence, capability, responsibility, rights of and trust in fathers to handle information, take parental leave, get parental allowance, be present in the life of the child, or be an unspecified role model (``boys/girls need male role models''). Note: If there are norms that fit the Caring or Daring categories and these are present in combination with ``other'' norms, these sentences will be annotated as Caring or Daring. In that way we prioritize caring and daring above the other category.

\paragraph{NOT APPLICABLE}
Sometimes sentences include the word father but do not tell us anything about fathers, and it is not possible to detect a norm. We asked the computer to denote those as ``not applicable.''

\paragraph{Additional labels}
The sentences are also labeled either as describing a state (DESCRIPTIVE) or an ideal (IDEAL). This is because we found that fatherhood norms can be either descriptive or prescriptive in the sentences (``we see that fathers are active'' or ``we wish for more active fathers''). Furthermore, fatherhood roles can be either IMPLICIT or EXPLICIT in the sentences. 

\section{Prompts}\label{app:prompts}

In this appendix, we provide the instructions and definitions of the labels included in our GPT3 prompts for each labeling dimension. Then, we provide the 15 examples used in few-shot learning. \\

{\noindent\bfseries Dimension 1: type of paternal involvement}

\noindent Instruction:
\begin{quote}
    Label the Swedish text according to how it describes the role of the father in the family.
    Possible labels are:
\end{quote}

\noindent Long labels definition: exactly the same reported in the previous appendix (\ref{app:corpus}).

\noindent Short labels definition:

\begin{quote}
    \begin{itemize}
    \item passive: fathers who are not actively involved in hands-on care and upbringing of the child;
    \item active\_negative: fathers exhibiting harmful behaviours like aggression, violence, or neglect;
    \item active\_positive\_caring: fathers providing care, warmth, empathy, and support;
    \item active\_positive\_challenging: fathers encouraging risk-taking, growth, and educational activities;
    \item active\_positive\_other: fathers displaying competence, responsibility, trustworthiness, etc., without specifying a specific role;
    \item not\_applicable: not applicable.  
    \end{itemize}
\end{quote}

{\noindent\bfseries Dimension 2: explicitness of the description}

\noindent Instruction:
\begin{quote}
    Label the Swedish text according to how it describes the role of the father in the family.
    Possible labels are:
    \begin{itemize}
    \item implicit
    \item explicit
    \item not\_applicable
    \end{itemize}
\end{quote}

\noindent In this case, we did not experiment with label definitions since labels are self explanatory.

{\noindent\bfseries Dimension 3: normativeness of the description}

\noindent Instruction:
\begin{quote}
    Label the Swedish text according to how it describes the role of the father in the family.
    Possible labels are: 
\end{quote}

\noindent Short labels definition:

\begin{quote}
    \begin{itemize}
    \item descriptive: if the sentence describes a state, like “we see that fathers are active”
    \item ideal: if the sentence is prescriptive, like “we wish for more active fathers”
    \item not\_applicable: not applicable.  
    \end{itemize}
\end{quote}

\noindent In this case, we did not experiment with longer label definitions since they would be redundant.

\subsection{Examples}\label{app:prompts:examples}

The examples used in few-shot learning were the following:

Text: i båda fallen är modern genetisk mor till barnet .
Label: not\_applicable

Text: i de fall socialnämnden inte tar sitt ansvar eller inte lyckas , måste det ändå finnas rätt för nära anhöriga , som mor- och farföräldrar , att hos tingsrätten föra talan om umgänge .
Label: not\_applicable

Text: det finns en överrepresentation bland fäder med låg utbildning , låg inkomst och pappor med utländsk bakgrund .
Label: passive, explicit, descriptive

Text: visserligen är det obalans i uttaget av föräldraförsäkringen , men samtidigt är det viktigt att slå fast att fler pappor i dag tar ut föräldraledighet än före införandet av mamma- och pappamånaden .
Label: active\_positive\_other, explicit, descriptive

Text: om barnets föräldrar varken är gifta eller registrerade samman- boende vid barnets födelse är mamman ensam vårdnadshavare .
Label: passive, implicit, descriptive

Text: bland dem fanns män , kvinnor och barn , unga och gamla , fäder och mödrar , söner och döttrar , makar och syskon .
Label: not\_applicable

Text: lagutskottet har tidigare anfört att man anser att den nuvarande ordningen ger betryggande garantier för att det verkligen är den biologiske fadern och inte någon annan som fastställs som far till ett barn ( 0000/00 : lu0 ) .
Label: passive, explicit, descriptive

Text: i isf : s rapport dubbeldagar – pappors väg in i föräldrapenningen dras slutsatsen att pappors uttag under barnets första levnadsår har ökat och att pappor som tidigare valt att ej ta ut föräldrapenning nu börjat ta ut dagar .
Label: active\_positive\_other, explicit, descriptive

Text: som förutsättning för alt barnavårdsnämnden skall vara skyldig att utreda om annan än den äkta mannen kan vara fader uppställer förslaget - utom att bamet skall ha viss anknytning lill sverige - dels att utredningen skall ha begärts av vårdnadshavare för barnet , d , v , s , normalt modern , eller av den äkta mannen , dels att det skall finnas lämpligt .
Label: passive, explicit, descriptive

Text: genom projektet `` läs för mig pappa '' , knutet till arbetsplatsbibliotek , uppmuntras fäder att både läsa själva och att läsa för sina barn .
Label: active\_positive\_challenging, explicit, ideal

Text: någon skyldighet att godkänna en utländsk dom om moderskap för barnets sociala eller genetiska mor , eller en utländsk dom om faderskap för en genetisk fars manliga partner , kan däremot inte utläsas av domstolens hittill- svarande praxis .
Label: passive, explicit, descriptive

Text: här uppmanas pappor också att ta ett ansvar som läsande fäder .
Label: active\_positive\_challenging, explicit, ideal

Text: av barnens svar kan man utläsa att mammor generellt har mer tid för sina barn oavsett i vilken utsträckning de bor med mamma , medan för barnen kan det ge mer tid med pappa när föräldrarna separerar än när de bor ihop under förutsättning att de minst bor halva tiden med pappan .
Label: active\_positive\_other, implicit, ideal

Text: det är fullt möjligt att kommunicera kring varför för- äldraskapet så lätt upplevs som något som angår mamman mer än pappan , varför pappor kan känna sig som tillräckligt goda fäder trots ett mycket lågt uttag av föräldraledighet , varför mammor ofta tror att de är bättre på att läsa av barnet än vad pappor är , varför pappor uppfattas som hjältar när de gör samma sak som mammor gör av bara farten och tusen andra frågor som har att göra med för- äldraskapet som bekönad position .
Label: active\_positive\_caring, implicit, ideal

Text: då utmålas oftast pappan som hotfull , våldsam och som en dålig förälder .
Label: active\_negative, explicit, descriptive

Text: det vanliga är sedan att våldtäktsmannen -- om det inte var fadern eller en nära släkting som förövat dådet -- betalar flickans föräldrar för att de inte skall anmäla fallet till polisen .
Label: active\_negative, implicit, descriptive

Text: för att underlätta för framför allt fäder att öka sitt uttag av föräldrapen­ning föreslås att ersättningen utges med sjukpenningens belopp oberoende av den andre förälderns sjukpenning .
Label: active\_positive\_other, explicit, ideal

Text: ha föräldrar gemensamt barn i sin värd , utgår föräldrapenning över garanlinivån lill fadern endast under förutsättning att även modern är eller enligt vad förut sagts bort vara försäkrad för sjukpenning som överstiger garantinivån .
Label: passive, explicit, descriptive

\section{Order of example sentences}\label{app:recency}
Not only does the number of examples that we provide in the prompt affect the output of a language model, also the order of the examples affect the output.
The last examples in our 15-shot sample have a more significant effect on the model predictions than earlier sentences. 
This known the ``recency effect'' \citep{zhao2021calibrate} and is based on how the language model was trained.
When predicting a missing word, the immediate context before the predicted word is more important than sentences that occurred several paragraphs before.
In the following, we study the effect of this recency bias on labelling performance and a strategy to deal with it. 
\newline

After constructing the prompt, we let GPT-3 label the validation set of 350 sentences by feeding it each example with the prompt via the API and collecting the answer label.
We used this setup to evaluate the performance of GPT-3 against human coders (see Section \ref{sec:eval}).
We evaluated different variations of the prompts to optimize for the best performances.
Once we were happy with GPT-3's performance, we let the LLM code the entire corpus, in our case 1,910 sentences.

\begin{table}[h]
\begin{tabular}{cccc}
\toprule
\textbf{Examples order} & \textbf{Kappa} & \textbf{Raw} & \textbf{F1} \\
\midrule
0 & \textbf{49.36} & \textbf{62.57} & \textbf{54.92} \\
1 & 48.27 & 61.90 & 53.51 \\
2 & 46.58 & 60.67 & 53.72 \\
\midrule
Avg & 47.97 & 61.71 & 54.05 \\
Std & 1.39 & 0.96 & 0.76 \\
\midrule
Majority & 48.40 & 61.90 & 54.31 \\
\bottomrule
\end{tabular}
\caption{GPT-3's performance on the first labeling task for three different random orders of the 15 labeled sentences provided in the prompt as examples.}\label{tab:exampleorder}
\end{table}

Table~\ref{tab:exampleorder} shows the results of three different GPT-3 runs over the validation set, where the prompt contains a different \textit{random} order of the 15 examples provided in each run.
The task under evaluation, in this case, is the first and most challenging (``type of paternal involvement''). We prompted the model with only the label list for this task and long, comprehensive descriptions.
As expected from previous findings on recency bias, the performance changes with the order of the examples, indicating that recency bias is at play here. 

Since we know which order performs best from trying different orders on the validation set, we can use this order when coding the entire corpus. In our case, order 0 exhibited the highest reliability vis-a-vis human coding.\footnote{Order 0 is also the one we used on the validation set in all the other experiments, as presented in the previous tables.}

Suppose we do not have a validation set available. In that case, we can mitigate the risk of recency bias by adopting a majority voting strategy.
In this case, we let the LLM code the whole corpus multiple times, with different random orders of the example sentences in the prompt, and then select for each instance the label that was output by the model most of the time, breaking ties randomly. Our results suggest that a majority label strategy is a viable option. It rendered a higher reliability (48.40 kappa) than the average (47.97) across the three individual runs. This result indicates we can minimize the risk of using an under-performing example order when a validation set is unavailable simply by running the LLM multiple times with different orders and then picking the majority label for each example returned by most runs. However, once again, the recommended approach is manually annotating a validation set and evaluating different random orders to maximize performance.


\section{Evaluation}\label{app:eval}
To evaluate the speed and cost of automatic coding, we let GPT-3\footnote{We choose GPT-3 over GPT-4 for this evaluation because it has higher rate limits. At the time of writing, GPT-4 was a brand-new technology, and its production was not yet mature. Therefore, GPT-3 was more representative of the speed and costs that the researchers will incur in the near future, when GPT-4 is expected to expand rate limits and lower costs.} annotate our entire dataset of 1,910 sentences with the best prompting strategy resulting from the previous evaluation step: 15-shot learning, providing a long, comprehensive description of the labels, and prompting the model to annotate across the three tasks jointly.

\subsection{Speed}\label{app:speed}
The completion time for GPT-3 was about 16 minutes.\footnote{This time can vary slightly depending on how busy OpenAI's API is.}
Meanwhile, based on other projects we have run, and ignoring diminishing returns due to fatigue, a trained and motivated human coder can label about 100 sentences of this type per hour. 
A single coder would thus take about 19 hours, or at least three work days, including the necessary breaks and excluding separate training sessions, adjudication of difficult cases, etc., to label our dataset.
If coding decisions became more complex, this time estimate would increase.

Thus not only does GPT-3, unlike other automatic coding approaches used in the literature, require no training time, it is also approximately 270 times faster than a human coder.

\subsection{Cost}
\label{app:cost}

The total price of a labeling request to OpenAI's API is computed based on the number of words in the prompt and response.\footnote{\url{https://openai.com/pricing}}
When coding a corpus like ours, the instruction, label definitions, and labeled examples present in the prompt are fixed across requests, while the sentence to be labeled changes at each API request.
Since the sentences to be coded are of different lengths, the price for coding each sentence varies.
The price of the response depends on the labels chosen by GPT-3, but it is negligible since each consists of only a few words.
In total, we spent \$93 coding the 1,910 sentences in our data,  an average price per instance of around 5 cents.

A realistic estimate for a human coder in the U.S. is a student assistant salary of \$12.35 per hour.\footnote{We use US standards for the cost of human coders since the pricing for GPT-3 is in US dollars}
Assuming the coder takes 19 hours to label the data, the total cost would be \$234.65.
That is more than 2.5 times the price of the LLM.
If we wanted multiple coders, as we simulated here, the overall price would go up, in our case, to a total of \$703.95.

In political science, we are often interested in country-specific text analysis, requiring native-speaking coders with appropriate knowledge of the field. In our case study with Swedish data, for example, this would incur a salary of about 170 SEK per hour, amounting to about \$881.79 for 19 hours of work for three coders, excluding recruitement costs.

As the LLM economy develops, APIs providing language modeling services are expected to become faster, cheaper, and more powerful, making LLMs an even more economic choice.

\subsection{Quality}
\label{app:quality}
Down below follows more evaluation metrics on the dataset. Besides average F1-score, the average Cohens kappa and average raw agreement across each evaluation run is presented. 

 \begin{table}[h]
 \begin{tabular}{lcccccc}
 \toprule
 \textbf{Annotator} & \textbf{Tasks} & \textbf{Label description} & \textbf{N. Examples} & \textbf{Kappa} & \textbf{Raw} & \textbf{F1} \\
 \midrule
 Human 1 & n.a. & n.a. & n.a. & 49.30 & 61.29 & 57.22 \\
 Human 2 & n.a. & n.a. & n.a. & 51.02 & 63.86 & 57.60 \\
 Human 3 & n.a. & n.a. & n.a. & 51.65 & 64.29 & 57.54 \\
 Humans AVG & n.a. & n.a. & n.a. & 50.66 & 63.15 & 57.45 \\
 \midrule
 GPT-3 & 1 & none & 0 & 13.41 & 27.71 & 22.73 \\
 GPT-3 & 1 & short & 0 & 20.89 & 36.38 & 34.27 \\
 GPT-3 & 1 & long & 0 & 18.06 & 33.90 & 34.69 \\
 GPT-3 & 1 & none & 15 & 36.53 & 51.05 & 44.22 \\
 GPT-3 & 1 & short & 15 & 44.50 & 58.48 & 49.40 \\
 GPT-3 & 1 & long & 15 & \textbf{49.36} & \textbf{62.57} & \textbf{54.92} \\
  GPT-3 & 1 & reasoned & 15 & 49.50 & 62.48 & 55.18 \\
 GPT-3 & 1,2,3 & long & 15 & 48.34 & 62.48 & 54.26 \\
 \midrule
 GPT-4 & 1,2,3 & long & 15 & \textbf{54.05} & \textbf{66.38} & \textbf{61.70} \\
  \midrule
  GPT-3 & 1 & long & 15 & 48.40 & 61.90 & 54.31 \\
  GPT-3 & 1 & reasoned & 15 & 48.93 & 62.19 & 54.49 \\
 \bottomrule
 \end{tabular}
 \caption{Average coding agreement with (other) human coders on task 1: type of paternal involvement (6 labels). We evaluate both human and GPT performance with different combinations of the prompt in terms of: annotation \textit{Tasks} performed simultaneously, length of the \textit{Label descriptions}, and \textit{N. of Examples} provided to the LLM.}\label{tab:main1}
 \end{table}

\begin{table}[h]
\begin{tabular}{lcccccc}
\toprule
\textbf{Annotator} & \textbf{Tasks} & \textbf{Label description} & \textbf{N. Examples} & \textbf{Kappa} & \textbf{Raw} & \textbf{F1} \\
\midrule
Human 1 & n.a. & n.a. & n.a. & 26.74 & 54.86 & 50.13 \\
Human 2 & n.a. & n.a. & n.a. & 26.55 & 58.29 & 48.61 \\
Human 3 & n.a. & n.a. & n.a. & 29.75 & 59.14 & 52.33 \\
Humans AVG & n.a. & n.a. & n.a. & 27.68 & 57.43 & 50.36 \\
\midrule
GPT-3 & 2 & none & 0 & 14.66 & 42.76 & 34.72 \\
GPT-3 & 2 & none & 15 & 22.91 & 53.14 & 48.54 \\
GPT-3 & 1,2,3 & none & 15 & \textbf{25.91} & \textbf{59.14} & \textbf{50.07} \\
\midrule
GPT-4 & 1,2,3 & none & 15 & \textbf{31.97} & \textbf{63.62} & \textbf{53.49} \\
\bottomrule
\end{tabular}
\caption{Average coding agreement with (other) human coders on task 2: explicitness of the description (2 labels).}\label{tab:main2}
\end{table}

\begin{table}[h!]
\begin{tabular}{lcccccc}
\toprule
\textbf{Annotator} & \textbf{Tasks} & \textbf{Label description} & \textbf{N. Examples} & \textbf{Kappa} & \textbf{Raw} & \textbf{F1} \\
\midrule
Human 1 & n.a. & n.a. & n.a. & 41.64 & 66.57 & 61.83 \\
Human 2 & n.a. & n.a. & n.a. & 37.00 & 67.14 & 57.94 \\
Human 3 & n.a. & n.a. & n.a. & 44.17 & 71.14 & 63.20 \\
Humans AVG & n.a. & n.a. & n.a. & 40.94 & 68.28 & 60.99 \\
\midrule
GPT-3 & 3 & none & 0 & 7.35 & 40.29 & 29.94 \\
GPT-3 & 3 & short & 0 & 7.96 & 36.86 & 32.00 \\
GPT-3 & 3 & none & 15 & 39.75 & 68.76 & 60.97 \\
GPT-3 & 3 & short & 15 & \textbf{44.08} & \textbf{70.86} & \textbf{64.25} \\
GPT-3 & 1,2,3 & short & 15 & 41.85 & 70.48 & 62.67 \\
\midrule
GPT-4 & 1,2,3 & short & 15 & \textbf{40.55} & \textbf{71.81} & \textbf{61.43} \\
\bottomrule
\end{tabular}
\caption{Average coding agreement with (other) human coders on task 3: normativeness of the description (2 labels).}\label{tab:main3}
\end{table}

\end{document}